\documentclass{article}
\title{Virtual Control Group: Measuring Hidden Performance Metrics}

\usepackage{hyperref}
\usepackage{authblk}
\author{Moshe Tocker}
\affil{Meta}

\date{June 2022}

\providecommand{\keywords}[1]
{
  \small	
  \textbf{\textit{Keywords---}} #1
}
\usepackage{graphicx}

\graphicspath{ {./images/} }
\raggedright

\usepackage{bbm}

\begin{document}

\maketitle

\section{ABSTRACT}

Performance metrics measuring in Financial Integrity systems are crucial for maintaining an efficient and cost effective operation. An important performance metric is False Positive Rate. This metric cannot be directly monitored since  we don't know for sure if a user is bad once blocked. We present a statistical method based on survey theory and causal inference methods to estimate the false positive rate of the system or a single blocking policy. We also suggest a new approach of outcome matching that in some cases including empirical data outperformed other commonly used methods.  The approaches described in this  paper can be applied in other Integrity domains such as Cyber Security. 
\linebreak

\keywords{causal inference, financial risk, outcome score matching, performance metrics, control group, false positive rate}

\section{INTRODUCTION}
Measuring performance metrics in Financial Integrity Systems  is crucial for decision making. 
Integrity systems need to hold a high level of precision and block bad users to reduce bad activity and comply with regulation and commercial requirements. On the other hand, risk systems that block many good users provide a terrible user experience which affects growth and adoption of the services. 

To monitor False Positives one can use a control group or periodically send transactions for manual review. Control Groups cost is high as they introduce  bad activity into the system. In addition, for Integrity systems in many cases this approach is not possible due to product or regulatory constraints. Manual labeling is another approach to estimate false positive rate. This method however requires manpower and  is extremely noisy (fraud labeling is a high-dimensional problem).

Another shortcoming of the approaches described above is measurement time delay. In the financial domain the most common indications of fraud are chargebacks. A chargeback is a return of money to a payer of a transaction, especially a credit card transaction after a successful dispute. This process takes time and therefore the indication of fraud takes time to arrive. This effect introduces a delay in the metrics measurement.

In this paper we present an alternative approach based on survey theory and causal inference methods that is purely statistical and has many advantages in terms of operational costs and time to measurement delay. We were able to predict the fraud prevalence among the blocked transactions  with acceptable errors in several use cases.

We believe that this approach can be useful in many integrity domains such as Financial Services, Cyber security  (Malware detection, Spam detection, …), Content Integrity and more.

\section{PRELIMINARIES}
Assume we have two functions of covariates vector x: Outcome Y(x) and Treatment assignment model W(x). In the financial integrity setting  we control the treatment assignment as we define the risk policies based on system risk signals (in contrast to the normal causal inference settings in which some treatment covariates are unknown). Units are sent to treatment based on the binary function W(x). W(x) can model a single fraud policy or the entire risk policies as a whole. The outcome function Y(x) is known only for units i of which W($x_i$)=0. We seek to estimate the mean value of the outcome of the treated group $E[Y|W=1]$. This problem can be viewed as  estimating population average of the non respondents from in a survey which has been dealt with by many authors such as \cite{robins1994estimation}  and \cite{hirano2001estimation}. In the setting of survey theory surveys are sent to a sampled population. Some of the people in the sample might not respond to the survey (a.k.a non-respondents) and thus the survey outcome would be unknown for them. Estimating the survey results just by the outcome of the respondents might lead to a biased result. We suggest that estimating false positive rate in financial risk is an equivalent problem in which the outcome is fraud or not-fraud and the non-respondents are the blocked transactions for which the outcome is unknown.

\subsection{Estimators}
\label{subsection:estimators}
The fields of Survey theory and causal inference produced statistical methods to try and replicate randomized experiments when only observational data is available. We describe in this section some of the methods of which we have considered. The method tries to estimate W(x), Y(x) or both based on the observed covariates and then draw conclusions over the mean of an unobserved outcome of the treated group. \linebreak \linebreak

Many estimation methods are based on the Propensity Score which is the probability of a unit to be treated and defined by \cite{robins1994estimation}. 
\begin{equation}
\pi(x)=P(W=1|X=x)
\end{equation}

Propensity Score  matching is a well-known approach used widely in practice. Each unit’s propensity score is estimated. Then units from the treated group are matched to  units from the untreated based on their nearest neighbor in terms of the propensity  score. The matching can be 1:1 also called pair matching or it can be k-nearest neighbor matching in which every item in the treated group is matched to k-nearest neighbors with replacement in the untreated group. Let $\mathcal{I}_{\pi}$ be the set of units in the untreated group which are the k-nearest neighbors by the propensity score of some unit in the treated group. Let $y_i$ be the true outcome of unit i. The set $\mathcal{I}_{\pi}$  is then used to estimate the mean outcome of the treated group:
\begin{equation} \label{eq:1}
    \hat{\mu}_{PSM}=\frac{1}{|\mathcal{I}_{\pi}|}\sum_{i \in \mathcal{I}_{\pi}}{y_i} 
\end{equation}

Let $\hat{m}(x)$ be a Logistic Regression model fitted on the outcome of the untreated group. Outcome score matching is done by matching units from the treated group  with units from the untreated group with nearest values of  $\hat{m}(x)$. Let $\mathcal{I}_{y}$ be the set of units in the untreated group which are the k-nearest neighbors by their predicted outcome of some unit in the treated group. The mean outcome is then estimating the mean the same way as in Equation \ref{eq:1}. 

\begin{equation}
    \hat{\mu}_{OSM}=\frac{1}{|\mathcal{I}_{y}|}\sum_{i \in \mathcal{I}_{y}}{y_i} 
\end{equation}

 We used matching on a single continuous covariate i,e the  outcome score since according to   \cite{abadie2006large} for the case when only a single continuous covariate is used to match,  the efficiency loss can be made arbitrarily close to zero by allowing a sufficiently large number of matches. 
 
 We have also considered inverse propensity weighting estimator  over the population of nonrespondents \cite{hirano2001estimation}
 
 \begin{equation}
     \hat{\mu}_{IPW-NR}=\frac{\sum_i{w_i{\pi}^{-1}(1-\pi)y_i}}{\sum_i{w_i\pi^{-1}(1-\pi)}}
 \end{equation}
 
 A different group  of estimators are regression estimators. These estimators model the outcome conditioned on the covariates using regression and use this model to estimate the missing outcomes. Let the  outcome model denoted by $\hat{m}(x)$ and $\mathcal{U}_t$ the set of  units in the treated group then the mean predicted outcome estimator is 
 
 \begin{equation}
\hat{\mu}_{MPO} = \frac{1}{|\mathcal{U}_t|}\sum_{i \in \mathcal{U}_t}{\hat{m}(x_i)}     
 \end{equation}

Doubly robust methods take into account both the model predictions for $y_i$ with inverse-probability weights. These methods are highly efficient when the y-model is true yet remain asymptotically unbiased when the y-model is misspecified \cite{kang2007demystifying}. We have considered such regression model for which the Logistic Regression was trained with non-respondents weights ${\pi}^{-1}(1-\pi)$  and refer to to it as $\hat{m}_w(x)$ so that the weighted mean predicted outcome estimator is :

 \begin{equation}
\hat{\mu}_{WMPO} = \frac{1}{|\mathcal{U}_t|}\sum_{i \in \mathcal{U}_t}{\hat{m}_w(x_i)}     
 \end{equation}

\subsection{confidence interval Estimation}
In contrast to  Machine Learning settings causal inference requires reporting of confidence intervals on top of the prediction. This is highly important so that the analyst can make a proper decision and know when prediction is of low quality. We use bootstrapping to estimate the standard error of all our estimators as recommended by many authors including \cite{austin2014use}. In bootstrapping  data variability estimation the dataset is sampled  many times with replacement to produce a sample with the same size. For each sample the estimator is being calculated. The standard deviation of the estimations is calculated and reported as the estimation of the standard error of the estimator. From this standard error (SE) the 95\% confidence interval can be simply derived by 

\begin{equation}
    CI 95\% = [\hat{\mu} - 1.96*SE , \hat{\mu} + 1.96*SE]
\end{equation}

\section{SIMULATION STUDIES}
\subsection{Treated Group Mean Estimation}
A simulation study was done to study the performance of different estimators for the potential mean outcome of the treated group. Design (1) is based on similar setting described by  \cite{pingel2018estimating}  and design (2) is similar to one described in \cite{austin2014use}

\begin{center} 
Treatment Design 1: $W(X)=\mathbbm{1}{[a_1X_1+a_2X_2+a_3X_3 > w ]}$ \linebreak
Outcome Design 1 : $Y(X)=\mathbbm{1}{[b_1X_1+b_2X_2+b_3X_3 + \epsilon > y]}$ \linebreak
Outcome Design 2 : $Y(X)=\mathbbm{1}{[b_1X_1^2+b_2X_2+b_3X_3 +\epsilon > y]}$ \linebreak
\linebreak
$X_i \sim \mathcal{N}(0,1) , a_i \sim \mathcal{U}(0,1), b_i \sim \mathcal{U}(0,1),\epsilon \sim \mathcal{N}(0,\sigma)$ 
\linebreak
$w \sim \mathcal{U}(-1,1),y \sim \mathcal{U}(-1,1)$
\end{center}

We use $\sigma$ to control the outcome model noise and thus the error rate of the estimation. For each design we sample  1000 data sets and estimate the mean outcome using all the estimators described in section \ref{subsection:estimators}.

The results are summarized in Table \ref{table:design1}, Table \ref{table:design2} and Table \ref{table:design2_100K}. All results in this paper are presented in percentage format (actual value times 100) for reader convenience. The method of Mean
predicted Outcome gave best results in our simulation studies in terms or Root mean square error and Mean Absolute error. We've noticed that an increase in the number of samples from 10K to 100K resulted in an improvement for the misspecified model but in quantities much lower than $\sqrt{n}$. The 10-NN outcome matching was the second runner up with low errors suggesting the benefits of matching over more than one unit. Methods not taking into account the outcome such as Inverse probability weighting and propensity score matching gave very bad results which makes them an unreasonable choice for the problem settings. 

\cite{kang2007demystifying} found  that the y-model and -model based  estimators are sensitive to misspecification models as not observing the direct confounders and rather observing some non-linear function of them. We have observed the same results for the methods tested. We’ve also noticed that introducing non-linearity by itself is enough to degrade the mean population estimation even if the covariates are observed. 

\begin{figure}
  \caption{Mean Estimation Vs True Mean for 6 different methods for Design 1 with 10K samples}
  \includegraphics[width=0.5\textwidth]{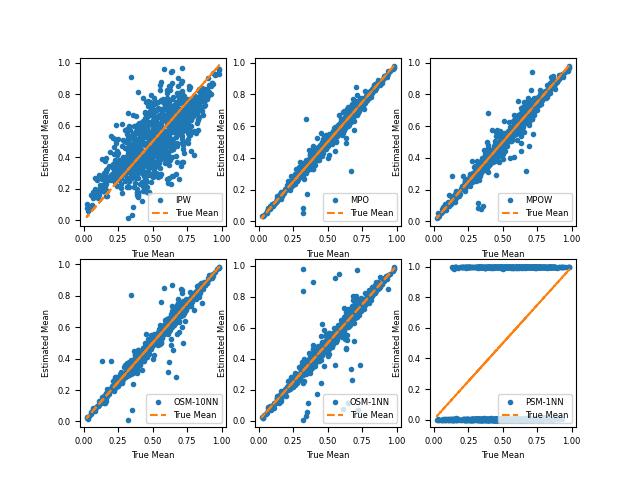}
\end{figure}

\begin{figure}
  \caption{Mean Estimation Vs True Mean for 6 different methods for Design 2 with 10K samples}
  \includegraphics[width=0.5\textwidth]{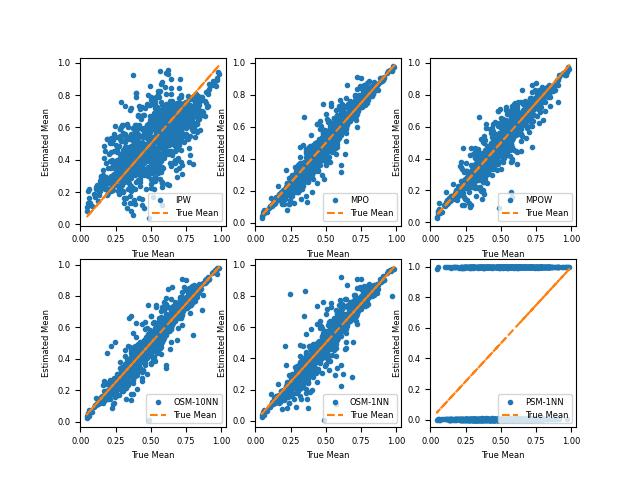}
\end{figure}

\begin{figure}
  \caption{Mean Estimation Vs True Mean for 6 different methods for Design 2 with 100K samples}
  \includegraphics[width=0.5\textwidth]{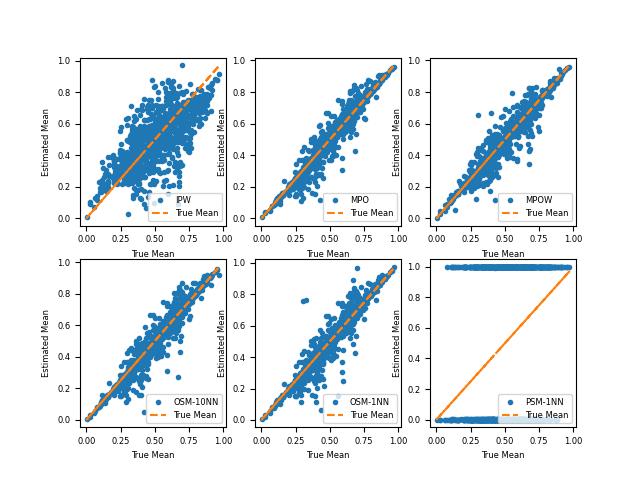}
\end{figure}

\begin{table} 

\caption{Performance of the estimators with 10,000 samples  over 1000 iterations for design 1 - correctly specified y-model and $\pi$-model 
\label{table:design1}}
\begin{tabular}{ l c c c } 
\hline
Method & BIAS & RMSE & MAE \\
\hline\hline
 Mean Predicted Outcome & 0.003 & 3.017 & 1.726 \\ 
 10-NN Outcome Matching & 0.077 & 4.09 & 2.059 \\ 
 IPW Mean Predicted Outcome & 0.111 & 3.879 & 2.268 \\ 
 1-NN Outcome Matching & 0.23 & 6.267 & 2.559 \\ 
 IPW & 0.353 & 13.512 & 10.714 \\
1-NN Propensity Score Matching & 2.088 & 48.309 & 44.327\\
 \hline
\end{tabular}
\end{table}

\begin{table} 
\caption{ Performance of the estimators with 10,000 samples  over 1000 iterations for design 2 - misspecified y-model and correctly specified  $\pi$-model 
\label{table:design2}
}
\begin{tabular}{ l c c c } 
\hline
Method & BIAS & RMSE & MAE \\
\hline\hline
Mean Predicted Outcome & -0.06 & 5.473 & 3.626 \\ 
10-NN Outcome Matching & -0.204 & 6.025 & 3.837 \\ 
1-NN Outcome Matching & 0.06 &7.128 & 4.21 \\ 
IPW Mean Predicted Outcome  &0.163 & 6.567 & 4.321 \\ 
IPW & 0.192 & 14.462 & 11.218 \\
1-NN Propensity Score Matching & 0.646 & 48.969& 45.304\\
 \hline
\end{tabular}
\end{table}

\begin{table} 
\caption{ Performance of the estimators with 100,000 samples  over 1000 iterations for design 2 - misspecified y-model and correctly specified 
$\pi$-model \label{table:design2_100K}
}
\begin{tabular}{ l c c c } 
\hline
Method & BIAS & RMSE & MAE \\
\hline\hline
Mean Predicted Outcome & 0.199 & 4.949 & 2.903 \\ 
10-NN Outcome Matching  & 0.359 & 5.192 & 2.95 \\ 
1-NN Outcome Matching  & 0.202 & 5.826 & 3.109 \\ 
IPW Mean Predicted Outcome & 0.206 &  5.477 & 3.21 \\ 
IPW & 0.242 & 13.699 & 10.612\\
1-NN Propensity Score Matching & -1.961 & 49.951 & 46.488 \\
 \hline
\end{tabular}
\end{table}

\subsection{Confidence Interval}
We've evaluated the confidence interval validity produced by bootstrapping. We ran the simulation again
with 100 samples of design 1. Each sample was of size 10K. Standard errors were estimated using 100 bootstrapped samples for each iteration. We have calculated the distribution ratio between the error and the estimated standard error and also the 95\% coverage rate presented in Tables \ref{table:design1_coverage}, \ref{table:design2_coverage}. For the correctly specified model all confidence intervals were estimated relatively good besides for the IPW confidence interval which underestimated the error with only about 24\% coverage. For the design 2 with misspecified outcome model our estimation of the confidence interval degraded with about ~75\% coverage. The only good estimation of the coverage rate was produced by the propensity score matching estimator.

\begin{table} 
\caption{ Design 1 Coverage Rate of different estimation methods using bootstrapping \label{table:design1_coverage}
}
\begin{tabular}{ l c c c } 
\hline
Method & Coverage Rate \\
\hline\hline
Mean Predicted Outcome & 94 \\
1-NN Outcome Score Matching &  96 \\
10-NN Outcome Score Matching & 92 \\
IPW Mean Predicted Outcome & 91 \\
1-NN Propensity Score Matching & 91 \\
IPW & 24 \\

 \hline
\end{tabular}
\end{table}

\begin{table} 
\caption{ Design 2 Coverage Rate of different estimation methods using bootstrapping \label{table:design2_coverage}
}
\begin{tabular}{ l c c c } 
\hline
Method & Coverage Rate \\
\hline\hline
1-NN Propensity Score Matching & 94 \\
1-NN Outcome Score Matching &  76 \\
10-NN Outcome Score Matching &  76 \\
IPW Mean Predicted Outcome & 75 \\
Mean Predicted Outcome & 72 \\
IPW & 23 \\

 \hline
\end{tabular}
\end{table}

\begin{figure}
  \caption{Design 1 Histograms of ratio between estimation error and the standard error}
  \includegraphics[width=0.5\textwidth]{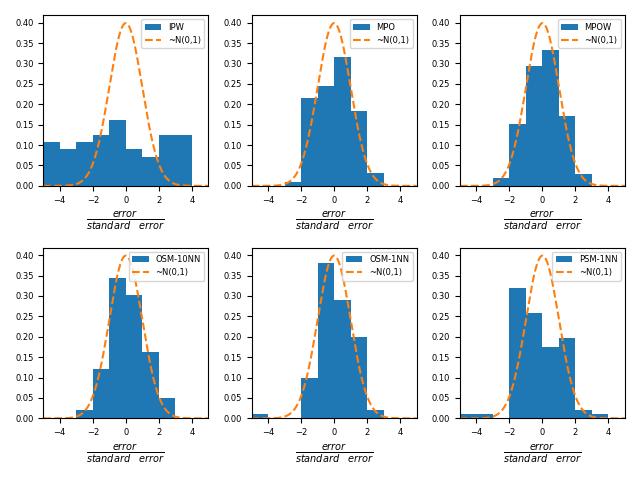}
\end{figure}

\begin{figure}
  \caption{Design 2 Histograms of ratio between estimation error and the standard error}
  \includegraphics[width=0.5\textwidth]{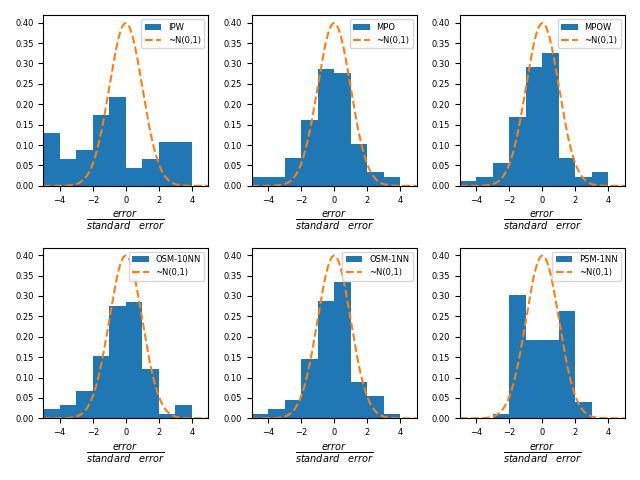}
\end{figure}

\section{EMPIRICAL RESULTS}
\subsection{Treated Group Mean Estimation} \label{subsection:treated group mean estimation}
To further investigate the estimator we’ve conducted an analysis over the public dataset “Credit Card Fraud Detection” from the Kaggle website. The dataset contains transactions made by credit cards in September 2013 by European cardholders. This dataset presents transactions that occurred in two days, with 492 frauds out of 284,807 transactions. The dataset is highly unbalanced, the positive class (frauds) account for 0.172\% of all transactions. It contains only numeric input variables which are the result of a PCA transformation. Due to confidentiality issues, the original features were not provided. Features V1, V2, … V28 are the principal components obtained with PCA, the only features which have not been transformed with PCA are 'Time' and 'Amount'. Feature 'Time' contains the seconds elapsed between each transaction and the first transaction in the dataset. The feature 'Amount' is the transaction Amount, Feature 'Class' is the response variable and it takes value 1 in case of fraud and 0 otherwise.

We performed feature selection using the Anova test and selected the top 10 features. To simulate a policy and treated group we sample at each iteration 4 features and create a mock policy using a Gauss Naive Bayes model. The data is split 50:50 to a train set and test set. The model is trained on the train set. The treated group size is fixed to 100 units. The treated group is assigned to the units with top model scores in the test set. Following that we run all the estimators and evaluate the results which are summarized in Table \ref{table:fraud_dataset}.

For the empirical results 10-NN score matching outperformed all other methods in terms of RMSE and MAE. IPW remains a bad choice for our domain but surprisingly 1-NN propensity score matching performance increased compared to our initial simulation studies. 

The empirical studies along with the simulation studies give strong evidence that these methods can be used in practice to get estimates of fraud prevalence and catch fraud trends.

\begin{table} 

\caption{ Performance of the estimators over 1000 iterations with Kaggle fraud  credit card dataset and treated group size of 100 \label{table:fraud_dataset}}
\begin{tabular}{ l c c c } 
\hline
Method & BIAS & RMSE & MAE \\
\hline\hline
10-NN Outcome Matching  & -1.77 & 3.82 & 3.26 \\ 
1-NN Outcome Matching  & -1.90 & 4.77 & 4.01 \\ 
Mean Predicted Outcome & -4.56 & 4.31 & 5.10 \\ 
IPW Mean Predicted Outcome & -5.81 &  7.25 & 7.63 \\ 
IPW & 7.37 & 6.97& 8.83 \\
1-NN Propensity Score Matching & 8.79 & 8.02 & 9.81 \\

 \hline
\end{tabular}
\end{table}

\begin{figure}
  \caption{Mean Estimation Vs True Mean for 6 different methods for Real Fraud Dataset from Kaggle website}
  \includegraphics[width=0.5\textwidth]{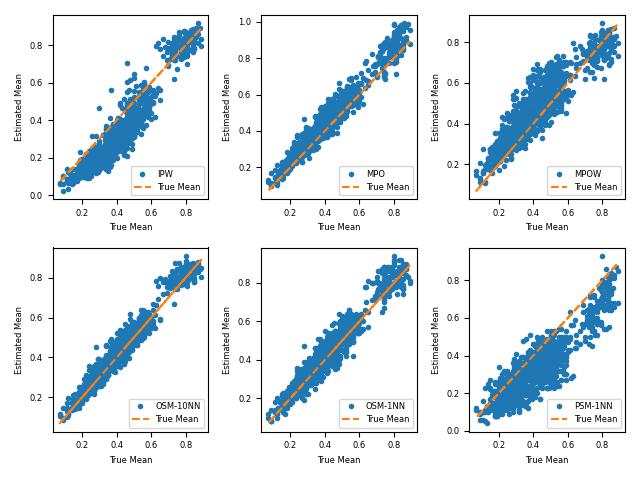}
\end{figure}

\subsection{Confidence Interval}
\begin{figure}
  \caption{Fraud Dataset Histograms of ratio between estimation error and the standard error 
  \label{figure:fraud_dataset_standard_error}}
  \includegraphics[width=0.5\textwidth]{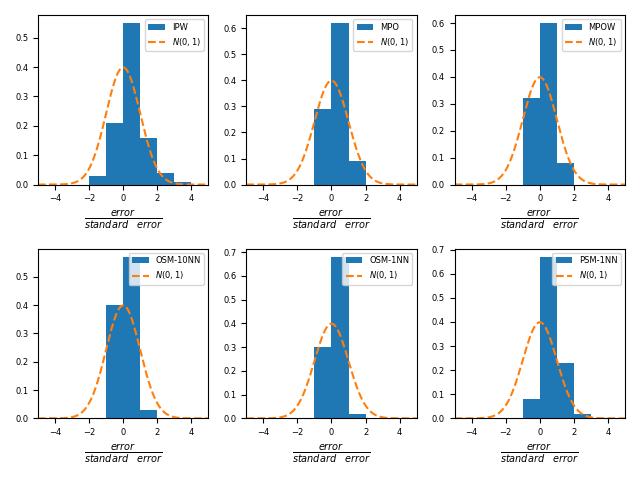}
\end{figure}

We evaluate the performance of bootstrapping in estimating the confidence interval of our estimators. We sample the Kaggle dataset 100 times for each estimation method. Each sample contains 900  non-fraud units and 100 fraud units.  We train a Gaussian Naive Bayse model on 10\% of the data with 4 random features. We then predict the model score over the entire set. We add random uniform noise $\mathcal{U}(0,1)$  to the Bayse Model score to get a higher variance in the positive ratio within the treated group. We select a random treated group size n with distribution $\mathcal{U}(50,80)$ and then assign the top n scores to the treated group. The results are summarized in Table \ref{table:fraud_dataset_coverage} and Figure \ref{figure:fraud_dataset_standard_error}.

\begin{table} 
\caption{ Kaggle Fraud Dataset Coverage Rate of different estimation methods using bootstrapping \label{table:fraud_dataset_coverage}
}
\begin{tabular}{ l c c c } 
\hline
Method & Coverage Rate \\
\hline\hline
IPW & 95 \\
1-NN Propensity Score Matching & 98 \\
1-NN Outcome Score Matching &  100 \\
10-NN Outcome Score Matching &  100 \\
IPW Mean Predicted Outcome & 100 \\
Mean Predicted Outcome & 100 \\

 \hline
\end{tabular}
\end{table}

It seems that bootstrapping for the use case of Kaggle fraud dataset with Naive Bayse policy overestimated the confidence interval and yielded conservative estimation.

\section{PRACTICAL CONSIDERATIONS}

\begin{enumerate}
  \item Null Ratio - special consideration should be given to the null ratio both feature level and item level. Since in production many times feature breakage comes in bursts and items with high null rate $(>10\%)$ will get the wrong match. We recommend using very reliable features and monitoring for breakage.

  \item Feature Selection - Features should be selected such that they explain both the outcome and the treatment. This can be verified using common unitary feature selection methods such as mutual information. Model should include as many covariates as possible to ensure correctness of the outcome model.

\end{enumerate}

\section{DISCUSSION} 
In this work we’ve presented a method for estimating unobserved performance metrics  with focus on Financial Services. We’ve shown how matching on the outcome score can predict metrics such as False Positive Rate with acceptable errors and serve as an additional tool. Moreover it was found to outperform other methods in a real fraud dataset. 

 We believe these methods, most of them known for many years in the statistics community, can be extremely useful in many Integrity domains for which control groups are not available and manual labeling effort reduction is desirable.

\section*{Acknowledgement}
The author would like to thank Shaked Bar for valuable comments.
\pagebreak
\bibliographystyle{plain}

\bibliography{ref.bib}
\nocite{*}

\end{document}